\documentclass{article}
\usepackage{spconf,amsmath,epsfig}
\usepackage{bbding}

\title{LIGHT: Joint Individual Building Extraction and Height Estimation \\
from Satellite Images through a Unified Multitask Learning Network}

\name{
Yongqiang Mao$^{1,2}$, Xian Sun$^{1,2}$, Xingliang Huang$^{1,2}$, Kaiqiang Chen$^{1,2,*}$\thanks{* Corresponding author.}
}
\address{$^{1}$Key Laboratory of Network Information System Technology (NIST), Aerospace \\
Information Research Institute, Chinese Academy of Sciences, Beijing, China\\
$^{2}$School of Electronic, Electrical and Communication Engineering, \\
University of Chinese Academy of Sciences, Beijing, China}

\begin{document}
\maketitle
\begin{abstract}
Building extraction and height estimation are two important basic tasks in remote sensing image interpretation, which are widely used in urban planning, real-world 3D construction and other fields.
Most of existing research regards the two tasks as independent studies. Therefore the height information cannot be fully used to improve the accuracy of building extraction, and vice verse.
In this work, we combine the individua\textbf{L} bu\textbf{I}lding extraction and hei\textbf{GH}t estimation through a unified multi\textbf{T}ask learning network (\textbf{LIGHT})
for the first time, which simultaneously outputs a height map, bounding boxes and a segmentation mask map of buildings. 
Specifically, \textbf{LIGHT} consists of an instance segmentation branch and a height estimation branch. 
In particular, so as to effectively unify multi-scale feature branches and alleviating feature spans between branches, we propose a Gated Cross Task Interaction (GCTI) module that can efficiently perform feature interaction between branches. Experiments on the DFC2023 dataset show that our \textbf{LIGHT} can achieve superior performance, and our GCTI module with ResNet 101 as the backbone can significantly improve the performance of multitask learning by 2.8\% AP50 and 6.5\% $\delta_1$, respectively.
\end{abstract}
\section{Introduction}
\label{sec:intro}
\begin{figure}
    \begin{center}
    \includegraphics[width=0.9\linewidth]{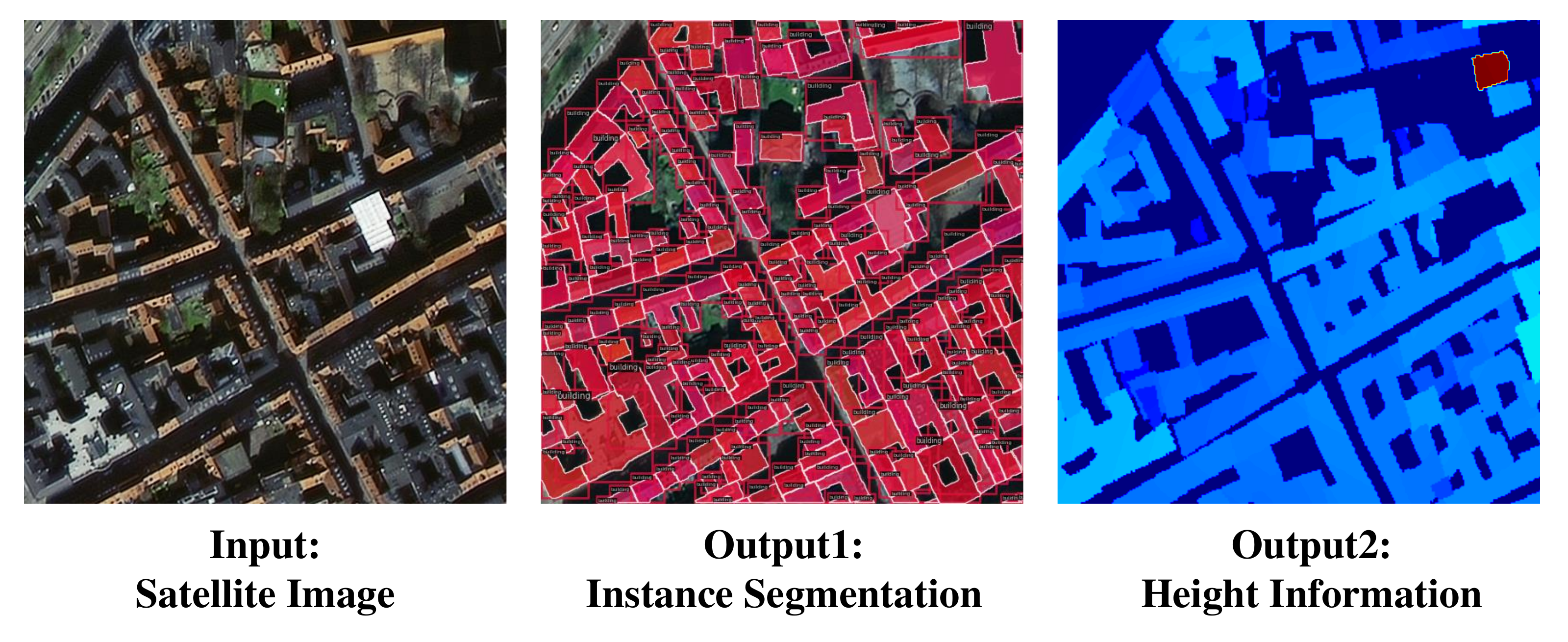}
    \end{center}
    \caption{Unified multitask learning for satellite imagery: Given (a) real orthophotos, the goal is to simultaneously output (b) instance segmentation results (including mask and bounding box) and (c) height information.}
    \label{motivation}
\end{figure}
Buildings play an essential role in urban activities. The acquisition and update of building layouts in cities are important for socioeconomic analysis, and urban planning. The technology of remote sensing and satellite image analysis provides a fast and cheap solution to building extraction and building height estimation in large scenes, which are the two key challenges. Besides, building extraction and height estimation play a pivotal role in 3D reconstruction of urban buildings and other fields.

\begin{figure*}
    \begin{center}
    \includegraphics[width=0.9\linewidth]{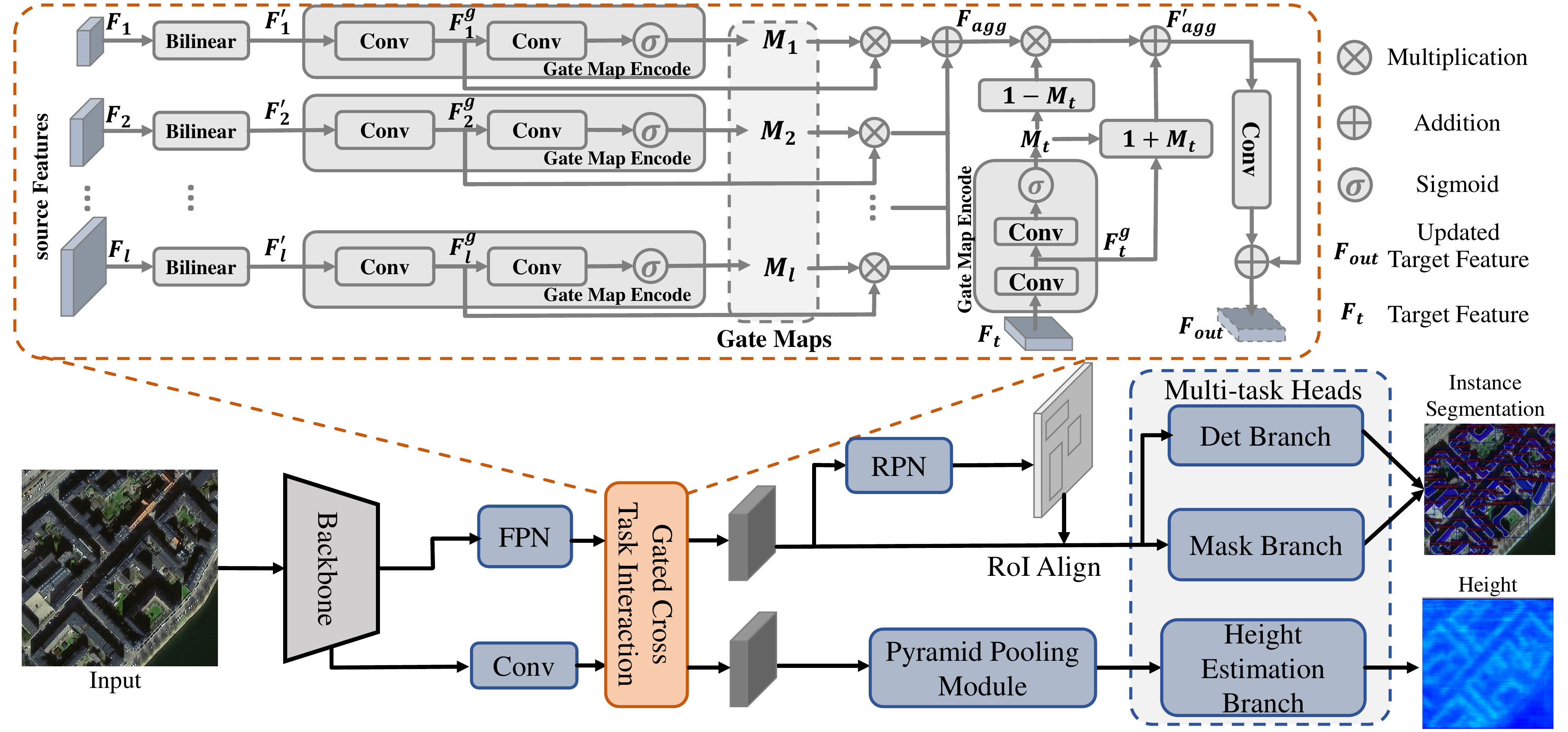}
    \end{center}
    \caption{The flowchart of our unified multitask learning network \textbf{LIGHT}. The input is a satellite image. The outputs include an instance segmentation result (including a mask map and bounding boxes) and a height estimation map.}
    \label{Framework}
\end{figure*}
The research on building extraction is mainly divided into pixel-wise extraction~\cite{chen2017building}, which focuses on pixel-wise building classification in images, and vector extraction~\cite{girard2020regularized}, which is to extract vectorial outlines of buildings by extracting key corner points of polygons. In addition, existing height estimation methods are mainly divided into discrete estimation~\cite{feng2022soft} and continuous estimation~\cite{mao2023elevation}, which produce discrete heights or continuous heights. Building3D~\cite{mao2023elevation} attempts to embed the two heterogeneous tasks into a framework to output pixel-wise building segmentation maps and height estimation maps, but it depends on two independent networks without considering the mutual feed of features. Although ASSEH~\cite{liu2022associatively} considers the mutual feed of cross-task features between networks, it cannot distinguish single individual building instances.


In this work, for joint individual building extraction and height estimation from satellite images, we propose a unified multitask Learning network named \textbf{LIGHT}. To the best of our knowledge, it is the first attempt to unifying these two independent tasks into a single end-to-end network. As shown in Fig. ~\ref{motivation}, our \textbf{LIGHT} can output a pixel-wise segmentation map, bounding boxes which are used to distinguish individual buildings, and a height map of buildings. Specifically, our \textbf{LIGHT} consists of a branch for building instance segmentation (including masks and bounding boxes) and a branch for height estimation of buildings. The two branches share a common backbone for feature extraction. The building instance segmentation branch is rooted in the Mask-RCNN~\cite{he2017mask} network. And the main components of the building height estimation branch are a Pyramid Pooling Module (PPM)~\cite{zhao2017pyramid} and a height estimation head. In addition, we propose a Gated Cross Task Interaction (GCTI) module, which promotes mutual information exchange between the two parallel branches, so as to improve the multitask learning ability of \textbf{LIGHT} network. Specifically, our GCTI module is composed of a basic gate unit, which aggregates features from two branches to different degrees by computing gate maps to achieve mutual feed of features. By introducing GCTI, our \textbf{LIGHT} is able to achieve efficient multitask learning by enhancing the feature communication between two branches. 


We conduct extensive experiments compared with the baseline. It demonstrates that our \textbf{LIGHT} has a significant enhancement over single-task learning methods, and demonstrates the effectiveness of our GCTI on multitask learning. On the DFC2023~\cite{huang2022urban} dataset, \textbf{LIGHT} can simultaneously achieve high-accuracy instance segmentation and height estimation.

\section{Proposed Method}
\label{sec:method}
In this paper, we propose a unified multitask learning network (\textbf{LIGHT}) for joint individual building extraction and height estimation, aiming to simultaneously output instance segmentation results (including masks and bounding boxes) and height estimation results, given a satellite image tile as input. 

\subsection{Unified Multitask Learning Network}
As shown in Fig. ~\ref{Framework}, our \textbf{LIGHT} network consists of two branches: an instance segmentation branch and a height estimation branch. The instance segmentation branch is further divided into two subbranches for bounding box and mask predictions, respectively. 
In the height estimation branch, \textbf{LIGHT} first extracts pyramidal features through a Pyramid Pooling Module (PPM)~\cite{zhao2017pyramid}, and then regresses the height of buildings through a height estimation head. The height estimation head is composed of a convolution layer, a batch normalization layer, a convolution layer, and a sigmoid operation in sequence, and finally outputs the pixel-wise height map, which can be formulated as:
\begin{equation}
    \mathbf{H} = Sigmoid(Conv(BN(Conv(\mathbf{F}))))
\end{equation}
where $Conv$ and $BN$ represents 2D convolution and batch normalization, respectively. $\mathbf{F}$ is the feature map derived from PPM. $\mathbf{H}$ is the output of the height estimation branch.

\textbf{Loss Function.} 
The loss function of our \textbf{LIGHT} framework consists of three parts: detection loss $\mathcal{L}_{det}$, mask loss $\mathcal{L}_{mask}$ and height loss $\mathcal{L}_{height}$. The detection loss and mask loss refer to MaskRCNN. And the height loss adopts Smooth L1 Loss. Then, the total loss $\mathcal{L}_{total}$ is obtained by combining three losses, which can be expressed as 
\begin{math}
    \mathcal{L}_{total} = \lambda_1 \mathcal{L}_{det} + \lambda_2 \mathcal{L}_{mask} + \lambda_3 \mathcal{L}_{height}
\end{math}, 
where $\lambda_1$, $\lambda_2$, and $\lambda_3$ are the weights. 
In the experiments, we set $\{\lambda_1=1.0, \lambda_2=1.0, \lambda_3=1.0\}$.

\begin{table*}[htb]
    \normalsize 
    \caption{The performance of \textbf{LIGHT} on DFC2023 dataset. HeightBase represents the height estimation branch of our \textbf{LIGHT} network. `w/ GCTI' means that \textbf{LIGHT} is embedding with Gated Cross Task Interaction (GCTI) module. 
    }\label{Performance}
    \centering
    \setlength\tabcolsep{9pt}
    \begin{tabular}{|l|c|c|cc|ccc|c|}
    \hline
    Method & w/ GCTI & Backbone & mAP$\uparrow$ & AP50$\uparrow$ & $\delta_1\uparrow$ & $\delta_2\uparrow$ & $\delta_3\uparrow$  & Inference time (ms)\\  
    \hline
    \hline
    MaskRCNN~\cite{he2017mask} &   -   & ResNet-50   & 22.8 & 49.1 & - & - & -  & 88 \\
    HeightBase  &   -   & ResNet-50   & - & - & 23.3 & 30.5 & 35.4 & 87 \\
    \textbf{LIGHT (ours)}      & \XSolidBrush & ResNet-50   & 22.2 & 53.9 & 38.2 & 43.2 & 47.4  & 108 \\
    \textbf{LIGHT (ours)}      & \Checkmark   & ResNet-50   & \bf{24.7} & \bf{57.4} & \bf{38.3} & \bf{44.0} & \bf{48.3}  & 127 \\
    \hline
    \hline
    MaskRCNN~\cite{he2017mask} &   -   & ResNet-101   & 23.2 & 55.2 & - & - & -  & 98 \\
    HeightBase  &   -   & ResNet-101   & - & - & 29.6 & 35.9 & 41.1 & 101 \\
    \textbf{LIGHT (ours)}      &   \XSolidBrush   & ResNet-101  & 22.5 & 54.9 & 44.6 & 48.6 & 52.3 &  115 \\
    \textbf{LIGHT (ours)}      &   \Checkmark     & ResNet-101  & \bf{25.3} & \bf{57.7} & \bf{51.1} & \bf{54.4} & \bf{57.6} & 137 \\
    \hline
    \end{tabular}
\end{table*}
\subsection{Gated Cross Task Interaction}
To enhance feature interactions between the multitask branches, we introduce a Gated Cross Task Interaction (GCTI) module. Through the GCTI module, features between the two branches achieve mutual feature enhancement. 
Specifically, our GCTI can make full use of the semantic information in the building extraction branch to enhance the learning ability of height features, and at the same time, the height of buildings can also benefit the extraction of buildings. 
Mathematically, $\{\mathbf{F}_{1}, \mathbf{F}_{2}, ...,\mathbf{F}_{l}\}$ are the semantic features derived from the FPN module, and $\mathbf{F}_h$ is the height feature derived from the convolution layer after the backbone in the height estimation branch. 
Select any of $\mathbf{F}_{i}$ and $\mathbf{F}_h$ as the target feature, and the rest as source features. Taking $\mathbf{F}_h$ as the target feature as an example, as $\mathbf{F}_t=\mathbf{F}_h$.
As shown in Fig. ~\ref{Framework}, firstly, the source features $\mathbf{F}_{i}$ are bilinearly interpolated to change the feature resolution to the target feature scale, as $\mathbf{F}^{'}_{i}=bilinear(\mathbf{F}_{i})$. Then, the source features $\mathbf{F}^{'}_{i}$ and target feature $\mathbf{F}_t$ are respectively passed through Gate Map Encode to generate gate maps. First, gate features $\mathbf{F}^{g}_{i}$ and $\mathbf{F}^{g}_{t}$ can be expressed as:
\begin{equation}
\begin{aligned}
      \{\mathbf{F}^{g}_{i}, \mathbf{F}^{g}_{t}\} = \{Conv_1(\mathbf{F}^{'}_{i}), Conv_1(\mathbf{F}_{t})\}\\    
\end{aligned}
\end{equation}
where $Conv_1$ is the 2D convolution. Next, we utilize 2D convolution and sigmoid operations to generate the source gate maps $\mathbf{M}_{i}$ and target gate map $\mathbf{M}_t$, as:
\begin{equation}    
      \{\mathbf{M}_{i}, \mathbf{M}_t\} = \{\sigma(Conv_2(\mathbf{F}^{g}_{i})), \sigma(Conv_2(\mathbf{F}^{g}_{t}))\}
\end{equation}  
where $Conv_2$ and $\sigma$ represents the convolution layer and the sigmoid operation, respectively. Next, the source gate map $\mathbf{M}_{i}$ is multiplied by the gate features $\mathbf{F}^{g}_{i}$, and then all source features are aggregated to $\mathbf{F}_{agg}$, which is expressed as:
\begin{equation}
    \mathbf{F}_{agg} = \sum^{l}_{i=1}\mathbf{M}_{i}\odot \mathbf{F}^{g}_{i}
\end{equation}
where $\odot $ is the Hadamard product. For the target feature branch, we set different gate weights to aggregate source and target gate features to $\mathbf{F}^{'}_{agg}$, which can be expressed as:
\begin{equation}
    \mathbf{F}^{'}_{agg} = (1+\mathbf{M}_t)\mathbf{F}^{g}_t+(1-\mathbf{M}_t)\mathbf{F}_{agg}
\end{equation}
Finally, the features $\mathbf{F}^{'}_{agg}$ are encoded into the final interaction feature, which can be expressed as 
$\mathbf{F}_{out} = Conv(\mathbf{F}^{'}_{agg})+\mathbf{F}^{'}_{agg}$. 
The processing of $\mathbf{F}_i$ as the target feature is the same.

\section{Experiments and Results}
\label{sec:results}
\subsection{Dataset and Inplementation}
\textbf{Dataset.} 
Experiments are conducted on the DFC2023~\cite{huang2022urban} track2 dataset, which consists of 12 cities across five continents. Furthermore, the dataset provides not only optical images but also synthetic aperture radar (SAR) images. The reference includes building annotations and a building height for each satellite image tile.


\textbf{Inplementation.} 
\textbf{LIGHT} is implemented on Pytorch framework and runs on NVIDIA RTX 3090 GPU. During training, the model is trained with a stochastic gradient descent (SGD) optimizer. Momentum is 0.9 and weight decay is 0.0001. For experiments, the image size is resized to 512$\times$512, and the total number of epochs, batch size and initial learning rate are set to 36, 2 and 0.02, respectively. 

\subsection{Quantitative Analysis}
\textbf{Performance.}
As shown in Table ~\ref{Performance}, we show the experimental performance of our \textbf{LIGHT} under different backbones along with ablation experiments.
First, whether ResNet50 or ResNet101 is chosen as the backbone, our unified multitask learning network is able to achieve performance comparable to or even significantly higher than that of single-task networks. 
This convincingly demonstrates the possibility and benefits of unifying the two heterogeneous tasks of building instance segmentation and height estimation in a single end-to-end network.
In particular, our unified multitask network (without GCTI) achieves 53.9\% AP50, 38.2\% $\delta_1$ when ResNet50 is used as the backbone, which are 4.8\% higher than MaskRCNN~\cite{he2017mask} and 14.9\% higher than HeightBase, respectively.
The introduction of GCTI can greatly improve the numerical performance of both tasks. In particular, when ResNet101 is used as the backbone, our \textbf{LIGHT} (with GCTI) achieves the best performance, reaching 57.7\% on AP50 and 51.1\% on $\delta_1$, which achieves 2.8\% and 6.5\% improvements on AP50 and $\delta_1$, respectively. This proves that GCTI can effectively learn height features and semantic features interactively, and using the features of different tasks can effectively enhance the performance of the heterogeneous tasks.

\textbf{Inference Time.} 
According to the inference time in the table, it is concluded that for the two single tasks to inference in turn, the inference speed of our \textbf{LIGHT} without GCTI (108 ms) is faster than the combination of two-task inference time (175ms: 88ms+87ms). Moreover, after adding GCTI, our inference time (127ms) is also better than the sum of the inference time of the two networks. 


\begin{figure}
    \begin{center}
    \includegraphics[width=1.0\linewidth]{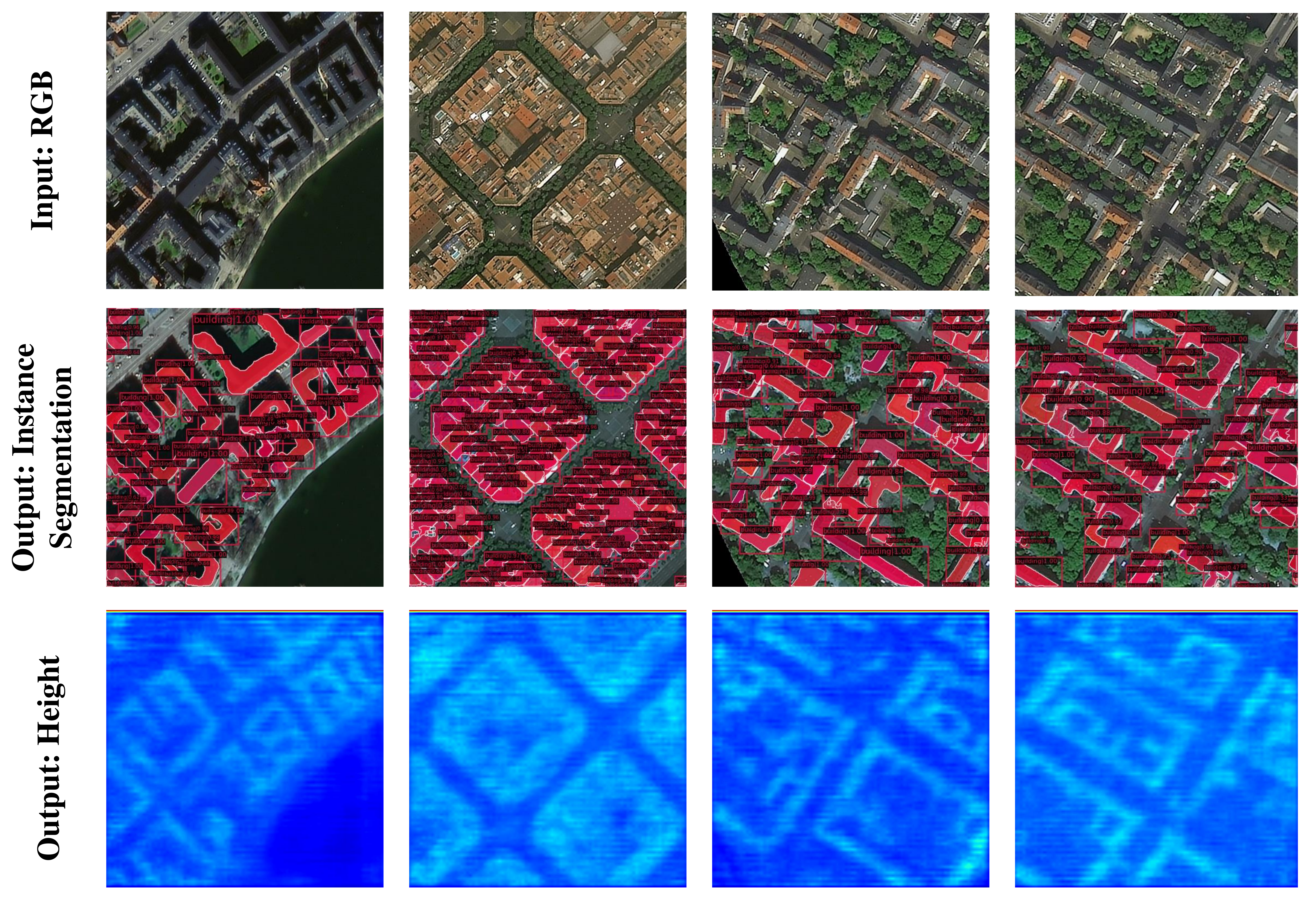}
    \end{center}
    \caption{Visualizations of our \textbf{LIGHT}. Input: RGB satellite image tiles (Line 1). Output: Instance segmentation (Line 2) and height estimation (Line 3) of buildings.}
    \label{Visual}
\end{figure}

\subsection{Qualitative Analysis}
As shown in Fig. ~\ref{Visual}, we present the visualization results of our method. 
The input is satellite image tiles, and the output is the instance segmentation (including mask and bounding box) and height estimation results of buildings. 
Qualitative visualization results reveal that our \textbf{LIGHT} achieve high-quality visualization results in building instance localization, segmentation, and height estimation. 
It demonstrates that our \textbf{LIGHT} is effective, showing that unifying instance segmentation and height estimation in one network is feasible.

\section{Conclusion}
\label{sec:conclusion}
In this paper, we propose a \textbf{LIGHT} network, which unifies instance segmentation and height estimation of buildings into a network for the first time. Further, we introduce GCTI to strengthen the feature mutual feed between the two branches. We conduct extensive experiments to demonstrate that \textbf{LIGHT} outperforms conventional single-task networks. In addition, the experiments prove that GCTI can effectively realize the efficient communication between cross-task features and improve the performance of both tasks. Our \textbf{LIGHT} provides new insights into multitask learning for joint individual building extraction and height estimation and remote sensing image analysis.


\bibliographystyle{IEEEbib}
\bibliography{refs}

\end{document}